\documentclass{article}

\usepackage[nonatbib,final]{neurips_2022_ml4ps}

\usepackage[utf8]{inputenc} 
\usepackage[T1]{fontenc}    
\usepackage{hyperref}       
\usepackage{url}            
\usepackage{booktabs}       
\usepackage{amsmath,amssymb,amsfonts}       
\usepackage{nicefrac}       
\usepackage{microtype}      
\usepackage{xcolor}         
\usepackage{graphicx}
\usepackage{mathrsfs}
\usepackage{amsthm}
\usepackage{subcaption}
\usepackage{algorithm}
\usepackage{comment}
\usepackage{algpseudocode}

\usepackage[style=ieee, citestyle=numeric-comp, backend=biber]{biblatex}
\definecolor{Blue}{rgb}{0,0,1}
\definecolor{Red}{rgb}{1,0,0}
\definecolor{Green}{rgb}{0,1,0}
\definecolor{Cyan}{rgb}{0,0.72,0.92}
\definecolor{Amethyst}{rgb}{0.6,0.4,0.8}
\definecolor{Bronze}{rgb}{0.8,0.5,0.2}
\definecolor{Violet}{rgb}{0.54,0.17,0.89}

\newcommand{\XH}[1]{{\textcolor{Red}{#1}}}

\title{Certified data-driven physics-informed greedy auto-encoder simulator}
\usepackage{multicol,multirow}
\renewcommand{\textcolor}[2]{#2} 

\author{
Xiaolong He \\
Department of Structural Engineering \\
University of California, San Diego \\
La Jolla, CA, 92093 \\
\texttt{xih251@ucsd.edu} \\
\And
Youngsoo Choi \\
Center for Applied Scientific Computing \\
Lawrence Livermore National Laboratory \\
Livermore, CA, 94550 \\
\texttt{choi15@llnl.gov} \\
\AND
William D. Fries \\
Applied Mathematics \\
School of Mathematical Sciences \\
University of Arizona \\
Tucson, AZ, 85721 \\
\texttt{frieswd@math.arizona.edu} \\
\And
Jonathan L. Belof \\
Materials Science Division \\
Physical and Life Science Directorate \\
Lawrence Livermore National Laboratory \\
Livermore, CA, 94550 \\
\texttt{belof1@llnl.gov} \\
\And
Jiun-Shyan Chen \\
Department of Structural Engineering \\
University of California, San Diego \\
La Jolla, CA, 92093 \\
\texttt{js-chen@ucsd.edu} \\
}

\begin{document}

\maketitle

\begin{abstract}
A parametric adaptive greedy Latent Space Dynamics Identification (gLaSDI) framework is developed for accurate, efficient, and certified data-driven physics-informed greedy auto-encoder simulators of high-dimensional nonlinear dynamical systems. 
In the proposed framework, an auto-encoder and dynamics identification models are trained interactively to discover intrinsic and simple latent-space dynamics. 
To effectively explore the parameter space for optimal model performance, an adaptive greedy sampling algorithm integrated with a physics-informed error indicator is introduced to search for optimal training samples on the fly, outperforming the conventional predefined uniform sampling.
Further, an efficient $k$-nearest neighbor convex interpolation scheme is employed to exploit local latent-space dynamics for improved predictability.
Numerical results demonstrate that the proposed method achieves \XH{121} to \XH{2,658} $\times$ speed-up with 1 to \XH{5}$\%$ relative errors for radial advection \XH{and 2D Burgers} dynamical problems.
\end{abstract}

\section{Introduction}\label{sec:introduction}
Physical simulations have played an increasingly significant role in the developments of engineering, science, and technology.
However, high-fidelity forward physical simulations can be computationally intractable even with high-performance computing, prohibiting their applications to problems that require a large number of forward simulations \cite{mcbane2021component,choi2020gradient,mcbane2022stress,wang2007large,white2020dual,choi2015practical,smith2013uncertainty,biros2011large,biros2011large,galbally2010non}.
In recent years, several reduced-order models (ROMs) have been integrated with latent-space learning algorithms \cite{kim2019deep,xie2019non,hoang2022projection,kadeethum2022non}. However, the latent-space dynamics models of these methods are often complex with limited \textit{interpretability}. 
Many methods have been developed for the identification of interpretable governing laws from data, including symbolic regression that searches both parameters and governing equations simultaneously \cite{koza1994genetic,schmidt2009distilling,cranmer2020discovering,cranmer2020pysr}, parametric models that fit parameters to equations of a given form, such as the sparse identification of nonlinear dynamics (SINDy) \cite{brunton2016discovering,champion2019data}, and operator inference \cite{peherstorfer2016data,qian2020lift,benner2020operator}.
Champion, et al. \cite{champion2019data} applied an auto-encoder for nonlinear projection and SINDy to identify simple ordinary differential equations (ODEs) that govern the latent-space dynamics. The auto-encoder and the SINDy model were trained interactively to achieve simple latent-space dynamics.
However, the proposed SINDy-autoencoder method is not parameterized and generalizable.
Many non-intrusive ROMs have been developed based on POD-based linear projection with latent-space dynamics captured by polynomials through operator inference \cite{peherstorfer2016data, geelen2022operator, guo2022bayesian, mcquarrie2021non, qian2020lift, benner2020operator}.
Due to the limitation of the POD-based linear projection, these non-intrusive ROMs have difficulties with advection-dominated problems.
To address this challenge, 
Fries et al. \cite{fries2022lasdi} proposed a parametric latent-space dynamics identification (LaSDI) framework, in which an auto-encoder was applied for nonlinear projection and a set of local dynamics identification (DI) models were introduced to identify local latent-space dynamics. 
The LaSDI framework can be viewed as a generalization of the aforementioned non-intrusive ROMs built upon latent-space dynamics identification since it allows linear or nonlinear projection and enables latent-space dynamics to be captured by flexible DI models based on general nonlinear functions.
However, the lack of interaction between the auto-encoder and the DI models due to a sequential training procedure leads to a strong dependency on the complexity and quality of the latent-space dynamics on the auto-encoder architecture, which could pose challenges to the subsequent dynamics identification and thus affect the model performances.
Most importantly, all these approaches rely on \textit{predefined} training samples, such as uniform or Latin hypercube sampling, which may not be optimal for the best model performance in the prescribed parameter space. 

To address these issues, we present a gLaSDI framework for accurate and efficient physics-informed data-driven reduced-order modeling. For an optimal sampling of the parameter space, an adaptive greedy sampling algorithm integrated with a physics-informed residual-based error indicator is introduced to search for optimal and minimal training samples on the fly. The proposed gLaSDI framework contains an auto-encoder for nonlinear projection to discover intrinsic latent representations and a set of local DI models to capture local latent-space dynamics, which is further exploited by an efficient $k$-nearest neighbors (k-NN) convex interpolation scheme.
The auto-encoder training and dynamics identification take place interactively to achieve an optimal identification of simple latent-space dynamics. Our numerical experiments demonstrate the effectiveness of the proposed gLaSDI framework and a considerable speed-up is achieved.

\section{Greedy Latent Space Dynamics Identification (gLaSDI)}\label{sec:glasdi}
\XH{Let us consider a system of ODEs characterizing a parameterized nonlinear dynamical system, which can also be considered as a semi-discretized equation of a system of partial differential equations (PDEs) with a spatial domain $\Omega$,}
\begin{equation}\label{eq.govern_eqn}
    \frac{d\mathbf{u}(t; \boldsymbol{\mu})}{dt} = \mathbf{f}(\mathbf{u},t; \boldsymbol{\mu}), \quad t \in [0,T], \quad \text{with} \quad \mathbf{u}(0; \boldsymbol{\mu}) = \mathbf{u}_0(\boldsymbol{\mu}),
\end{equation}
where $T$ is the final time; 
$\boldsymbol{\mu}$ is the parameter in a parameter domain $\mathcal{D}$;
$\mathbf{u} \in \mathbb{R}^{N_u}$ is the parameterized time-dependent solution to the dynamical system with an initial state $\mathbf{u}_0$;
$\mathbf{f}$ denotes the velocity.
With the implicit backward Euler time integrator, \XH{an approximate solution} to Eq. (\ref{eq.govern_eqn}) can be obtained by solving the nonlinear system of equations $\mathbf{u}_n = \mathbf{u}_{n-1} + \Delta t \mathbf{f}_n$,
where $\mathbf{u}_n := \mathbf{u}(t^n; \boldsymbol{\mu})$, and $\mathbf{f}_n := \mathbf{f}(\mathbf{u}(t^n; \boldsymbol{\mu}), t^n; \boldsymbol{\mu})$. 
The residual function \XH{is defined as} 
\begin{equation}\label{eq.residual}
    \mathbf{r}(\mathbf{u}_n; \mathbf{u}_{n-1}, \boldsymbol{\mu}) = \mathbf{u}_n - \mathbf{u}_{n-1} - \Delta t \mathbf{f}_n.
\end{equation}

The physics-informed adaptive greedy sampling is performed on a discrete parameter space ($\mathcal{D}^h \subseteq \mathcal{D}$) and 
$\mathbb{D} \subseteq \mathcal{D}^h$ denotes a set of $N_{\mu}$ selected training sample points. 
Let's consider the $i$-th training sample point $\boldsymbol{\mu}^{(i)} \in \mathcal{D}^h$ and $\mathbf{u}_n^{(i)}$ as the associated solution at the $n$-th time step of the dynamical system in Eq. (\ref{eq.govern_eqn}). 
The solutions at all time steps are arranged in a snapshot matrix denoted as $\mathbf{U}^{(i)} = [\mathbf{u}_0^{(i)}, ..., \mathbf{u}_{N_t}^{(i)}]$. 
The \textit{reconstruction loss} of the auto-encoder is defined as $\mathcal{L}_{recon} := || \mathbf{U} - \hat{\mathbf{U}} ||_{L_2}^2$, 
where \XH{ $\mathbf{U} = \{ \mathbf{U}^{(j)}\}_{j=1}^{N_{\mu}}$ and $\hat{\mathbf{U}} = \{ \hat{\mathbf{U}}^{(j)} \}_{j=1}^{N_{\mu}}$ },
$\hat{\mathbf{U}}^{(i)} = [\hat{\mathbf{u}}_0^{(i)}, ..., \hat{\mathbf{u}}_{N_t}^{(i)}]$, 
$\hat{\mathbf{u}}_n^{(i)} = \boldsymbol{\phi}_d(\mathbf{z}_n^{(i)})$, 
and $\mathbf{z}_n^{(i)} = \boldsymbol{\phi}_e(\mathbf{u}_n^{(i)})$; 
$\boldsymbol{\phi}_e$ is an encoder 
and $\boldsymbol{\phi}_d$ is a decoder;
$\mathbf{z}_n^{(i)} \in \mathbb{R}^{N_z}$ denotes the latent variable with $N_z \ll N_u$.

Given the latent variable in the discrete form 
$\mathbf{Z}^{(i)} = [\mathbf{z}_0^{(i)}, ..., \mathbf{z}_{N_t}^{(i)}]$ obtained from the encoder, the governing equation of its dynamics is approximated by a DI model, $\boldsymbol{\psi}_{DI}$, based on a user-defined library of basis functions $\boldsymbol{\Theta}(\mathbf{Z}^{(i)T})$, e.g., polynomial, trigonometric, and exponential functions, such that $\dot{\mathbf{Z}}^{(i)T} \approx \dot{\hat{\mathbf{Z}}}^{(i)T} = \boldsymbol{\Theta}(\mathbf{Z}^{(i)T}) \boldsymbol{\Xi}^{(i)}$,
where $\boldsymbol{\Xi}^{(i)}$ is a coefficient matrix. See \cite{fries2022lasdi} for more details. 

\begin{figure}[htp]
    \centering
    \includegraphics[width=1.0\textwidth]{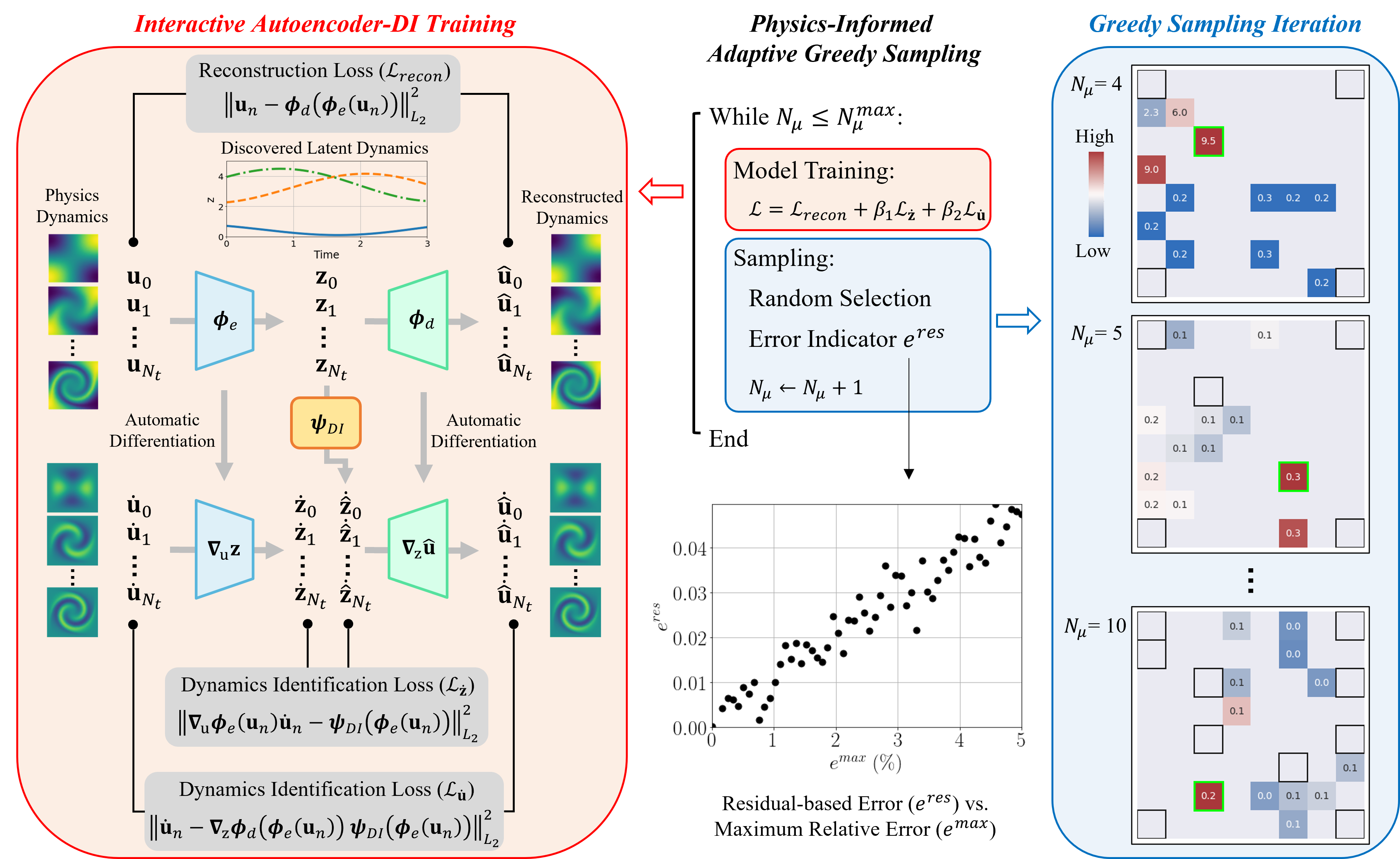}
    \caption{Schematics of the gLaSDI algorithm. \XH{The black square boxes in the right panel indicate the sampled parameter points in a 2D parameter space.}}
    \label{fig.glasdi}
\end{figure}

To identify simple and smooth latent-space dynamics, two additional loss functions, $\mathcal{L}_{\dot{\mathbf{z}}}$ and $\mathcal{L}_{\dot{\mathbf{u}}}$, are constructed to enable interactive training of the auto-encoder and the DI model (\textit{interactive Autoencoder-DI training}), as illustrated in Fig. \ref{fig.glasdi}, 
where $\mathcal{L}_{\dot{\mathbf{z}}} := || \dot{\mathbf{Z}} - \dot{\hat{\mathbf{Z}}} ||_{L_2}^2$ ensures the consistency of the \textit{latent-space dynamics gradients}, 
and $\mathcal{L}_{\dot{\mathbf{u}}} := || \dot{\mathbf{U}} - \dot{\hat{\mathbf{U}}} ||_{L_2}^2$ ensures the consistency of the \textit{physical dynamics gradients}.
Note that $\dot{\mathbf{Z}}$ and $\dot{\hat{\mathbf{U}}}$ can be computed through the chain rule, as illustrated in Fig. \ref{fig.glasdi}. 
The local DI models are considered to be point-wise (see more details in \cite{fries2022lasdi}), which means each local DI model is associated with a distinct sampling point in the parameter space. Hence, each sampling point has an associated DI coefficient matrix.

\XH{It is expected that the latent-space dynamics within a small neighborhood of a parameter point are similar.} 
To exploit the local latent-space dynamics captured by the local DI models for enhanced parameterization and efficiency, a k-NN convexity-preserving partition-of-unity interpolation scheme based on Shepard function \cite{shepard1968two} or inverse distance weighting is employed.
The associated DI coefficient matrix of a testing parameter $\boldsymbol{\mu} \in \mathcal{D}$ is obtained by a convex interpolation of coefficient matrices of its $k$-nearest neighbors (existing sampling points), $\boldsymbol{\Xi}_{interp} = \sum_{i \in \mathcal{N}_k(\boldsymbol{\mu})} \Psi^{(i)}(\boldsymbol{\mu}) \boldsymbol{\Xi}^{(i)}$.
The $k$-nearest neighbors are selected based on the \XH{Mahalanobis} distance between the testing and the training (existing) parameters. 
The interpolation functions $\Psi^{(i)}(\boldsymbol{\mu})$ with an inverse-distance kernel function is employed, preserving convexity and a partition of unity. 

To address the issues of parameter dependency of local latent-space dynamics efficiently and effectively, a physics-informed adaptive greedy sampling procedure is introduced to construct an optimal training database $\mathcal{DB}=\{\mathbf{U}^{(i)}\}_{i=1}^{N_{\mu}}$ on the fly during off-line interactive autoencoder-DI training, as illustrated in Fig. \ref{fig.glasdi}, which corresponds to a set of parameters $\mathbb{D}=\{\boldsymbol{\mu}^{(i)}\}_{i=1}^{N_{\mu}}$ sampled from $\mathcal{D}^h$ with $N_{\mu} < = |\mathcal{D}^h|$.
The training database is first initialized with a small set of parameters located, e.g., at the corners or at the center of the parameter space. 
To enhance sampling reliability and quality, the model training is performed for a number of epochs before every greedy sampling, as illustrated in Fig. \ref{fig.glasdi}. 
To expedite the sampling procedure, a random-subset evaluation strategy is adopted.
At the $v$-th sampling iteration, a small set of candidate parameters, $\mathbb{D}^{sub} \subseteq \mathcal{D}^h$ ($\mathbb{D}^{sub} \cap \mathbb{D}_{v-1} = \emptyset$), are randomly selected and the parameter $\boldsymbol{\mu}^*$ that maximizes a \textit{residual-based error indicator}, $e^{res}(\boldsymbol{\mu})$, is selected \XH{to update}
$\mathbb{D}_v = \{\mathbb{D}_{v-1}, \boldsymbol{\mu}^* \}$ and 
$\mathcal{DB}_v = \{\mathcal{DB}_{v-1}, \mathbf{U}^{*} \}$.
The residual error indicator is defined as 
\begin{equation}\label{eq.error_indicator}
    e^{res} = \frac{1}{N_{ts}+1} \sum_{n=0}^{N_{ts}}||\mathbf{r}(\hat{\mathbf{u}}_n; \hat{\mathbf{u}}_{n-1}, \boldsymbol{\mu})||_{L_2},
\end{equation}
\XH{where the residual $\mathbf{r}(\hat{\mathbf{u}}_n; \hat{\mathbf{u}}_{n-1}, \boldsymbol{\mu})$ is physics-informed and obtained from Eq. \eqref{eq.residual} that relies on discretized governing equations;} $N_{ts} \approx 0.1 N_t$ to enhance the computational efficiency. 
The residual error indicator is \XH{calculated based on only predictions and positively correlated with the maximum relative error}, as demonstrated in Fig. \ref{fig.glasdi}. 
\XH{The maximum residual error ($e_v^{res}$) of the training samples at the $v$-th iteration can be used to estimate the maximum relative error ($e_v^{max}$) in the parameter space based on a linear least-square fit between $\mathbf{E}_v^{res} = \{ e^{res} \big( \hat{\mathbf{U}}(\boldsymbol{\mu}) \big) \}_{\boldsymbol{\mu} \in \mathbb{D}_v}$ and 
$\mathbf{E}_v^{max} = \{ e^{max} \big( \mathbf{U}(\boldsymbol{\mu}), \hat{\mathbf{U}}(\boldsymbol{\mu}) \big) \}_{\boldsymbol{\mu} \in \mathbb{D}_v}$.}
The greedy sampling procedure continues until a prescribed number of sampled points \XH{or a target error tolerance} is reached.
The trained gLaSDI model can then be applied to efficiently predict dynamical solutions given a testing parameter \XH{in the prescribed parameter space}.

\section{Results and Discussion}\label{sec:result}
The performance of gLaSDI is demonstrated by solving a two-dimensional parameterized radial advection problem and compared with that of LaSDI \cite{fries2022lasdi} (without \XH{autoencoder-DI interactive training and} adaptive greedy sampling):
\begin{equation}\label{eq.advection}
    \frac{\partial u}{\partial t} + \mathbf{v} \cdot \nabla  u = 0, \quad \Omega = [-1,1]\times[-1,1], \quad t \in [0,3],
\end{equation}
with a boundary condition $u(\mathbf{x},t;\boldsymbol{\mu}) = 0$ on $\partial\Omega$ and an initial condition $ u(\mathbf{x},0;\boldsymbol{\mu}) = \text{sin}(w_1 x_1) \text{sin}(w_2 x_2)$ parameterized by $\boldsymbol{\mu} = \{w_1, w_2\}$,
where $\mathbf{v} = \frac{\pi}{2} d [x_2, -x_1]^T$ denotes the fluid velocity with $d = (1-x_1^2)^2(1-x_2^2)^2$. 
The spatial domain is discretized by first-order periodic square finite elements constructed on a uniform grid of $96 \times 96$ discrete points. 
The fourth-order Runge-Kutta explicit time integrator with a uniform time step of $\Delta t = 0.01$ is employed. 
A discrete parameter space $\mathcal{D}^h$ is constituted by the parameters of the initial condition, including \XH{$w_1 \in [1.5, 1.8]$ and $w_2 \in [2.0, 2.3]$}, each with 21 evenly distributed discrete points in the respective parameter range.
The distributions of the solution field and its gradient at a few time steps for the parameter case $(w_1=1.5, w_2=2.0)$ are shown in Fig. \ref{fig.glasdi}.
\XH{The gLaSDI model consits of \textit{Linear} DI models and an auto-encoder that has} an architecture of 9,216-100-3-100-9,216\XH{, with the numbers denoting the number of neurons in each hidden layer, and the latent dimension as 3}. 
The training and testing are performed on a NVIDIA V100 (Volta) GPU 
with 3,168 NVIDIA CUDA Cores and 64 GB GDDR5 GPU Memory. The open-source TensorFlow library \cite{abadi2016tensorflow} and the Adam optimizer \cite{kingma2014adam} are employed for model training. 
The gLaSDI training is performed until the total number of sampled parameter points reaches 25. 
A LaSDI model with the same architecture of the auto-encoder and DI models is trained using 25 predefined training points uniformly distributed in a $5 \times 5$ grid in the parameter space.

\begin{figure}[htp]
    \centering
    \includegraphics[width=1.0\textwidth]{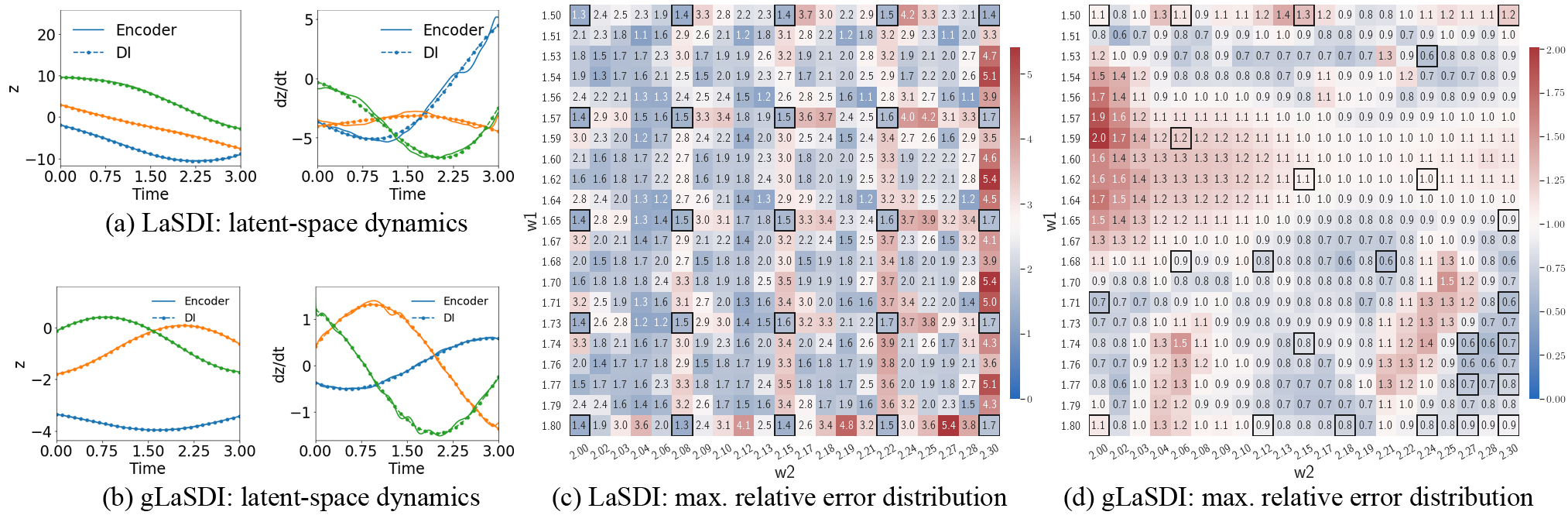}
    \caption{Comparison between LaSDI and gLaSDI with the same model architecture for the radial advection problem with parameters $w_1 \in [1.5, 1.8]$ and $w_2 \in [2.0, 2.3]$. The number on each box denotes the maximum relative error of the associated parameter case. The black square boxes indicate the sampled training parameter points.}
    \label{fig.advection_case1}
\end{figure}

Owning to the interactive autoencoder-DI training, gLaSDI is able to capture simpler and smoother latent-space dynamics than LaSDI that has sequential and decoupled training of the auto-encoder and DI models, as shown in Figs. \ref{fig.advection_case1}(a-b). 
\XH{A better agreement between the latent-space dynamics predicted by the encoder and the DI models is achieved by gLaSDI.
Figs. \ref{fig.advection_case1}(c-d) show that gLaSDI achieves a maximum relative error of 2.0$\%$ in the whole parameter space, lower than 5.4$\%$ of LaSDI.}

\begin{figure}[htp]
    \centering
    \includegraphics[width=1.0\textwidth]{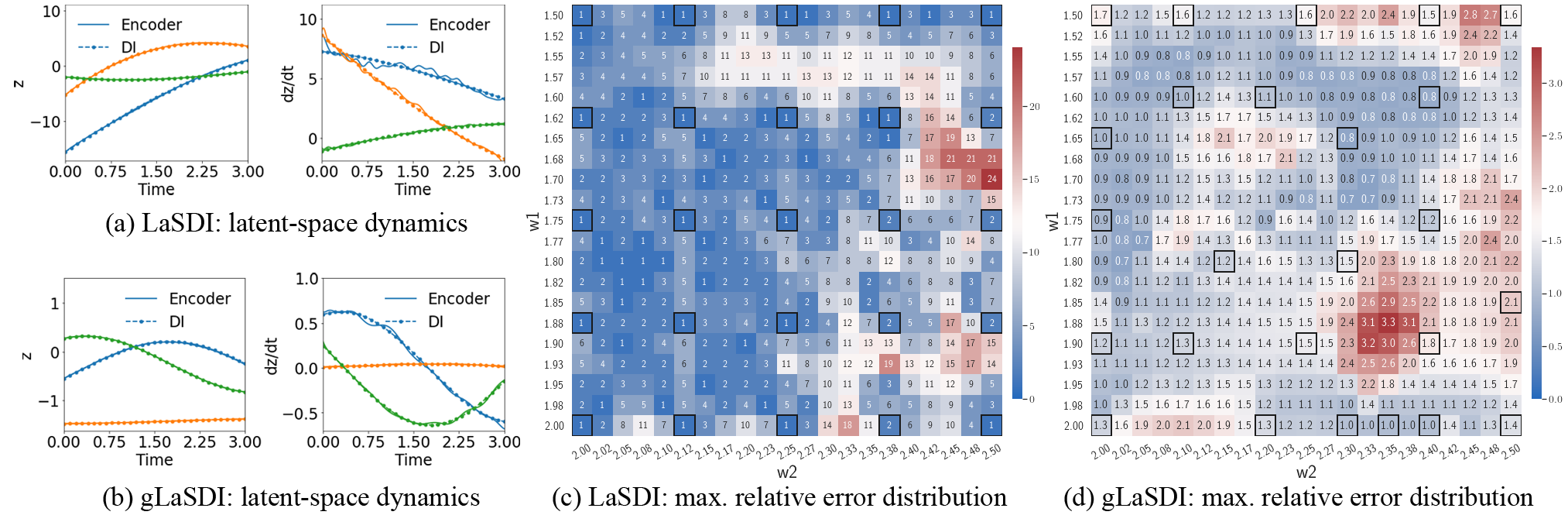}
    \caption{Comparison between LaSDI and gLaSDI with the same model architecture for the radial advection problem with parameters $w_1 \in [1.5, 2.0]$ and $w_2 \in [2.0, 2.5]$. The number on each box denotes the maximum relative error of the associated parameter case. The black square boxes indicate the sampled training parameter points.}
    \label{fig.advection_case2}
\end{figure}

\XH{When the parameter space is enlarged to $w_1 \in [1.5, 2.0]$ and $w_2 \in [2.0, 2.5]$, gLaSDI maintains a high accuracy with a maximum relative error of 3.3$\%$, much lower than 24$\%$ of LaSDI, as shown in Figs. \ref{fig.advection_case2}(c-d).}
The results demonstrate that given the same number of training parameters, gLaSDI with the physics-informed adaptive and sparse sampling can intelligently identify the optimal training parameter points to achieve higher accuracy than LaSDI based on predefined uniformly distributed training parameters.
\XH{It is interesting to note that changing the parameter space affects the distribution of
gLaSDI sampling.}
Compared with the high-fidelity simulation based on MFEM \cite{anderson2021mfem}, the gLaSDI model achieves \XH{121}$\times$ speed-up.

A potential limitation is the residual computation, which might not be straight-forward, for example, if the source code of the full-order model is not available. Then, the physics information needs to be post-processed with the gLaSDI solution, which might not be accurate. 
Another limitation is that the training of the auto-encoder can be computationally expensive as the data size becomes large\XH{, which can be improved by applying the convolutional auto-encoder.}
\XH{The training efficiency can potentially be further improved by a combination of \textit{pre-training} and \textit{re-training}}. 
The proposed gLaSDI framework is general and not restricted by the use of auto-encoders and DI models.
Depending on applications, linear compression techniques with much lower training cost could be employed. 
Further, other system identification techniques or operator learning algorithms could be applied to identify latent-space dynamics.
The auto-encoder architecture can be optimized to maximize generalization performance by integrating automatic neural architecture search into the proposed framework.
The parameterization in this study only considers the parameters from the initial conditions of the problems. 
The proposed framework can be easily extended to account for other parameterization types, such as of material properties, which will be useful for inverse problems. 
All the codes to regenerate the results in this paper can be found in the GitHub repository
(\url{https://github.com/LLNL/libROM/tree/gLaSDI/python/gLaSDI}).
The following existing assets have been used: SindyAutoencoder with MIT license (\url{https://github.com/kpchamp/SindyAutoencoders}) and the MFEM library with BSD-3-Clause (\url{https://github.com/mfem/mfem}).

\newpage
\section{Broader impact}\label{sec:impact}
This paper introduces a novel machine learning enhanced physics-guided data-driven reduced-order modeling strategy with adaptive sampling to accelerate high-dimensional physical simulations. 
The proposed gLaSDI framework is expected to have broad impacts on the computational science community and application potentials in a wide range of engineering and scientific domains.
There is no negative consequence on ethics and society in this work.

\section*{Appendix}
\begin{figure}[htp]
    \centering
    \includegraphics[width=1.0\textwidth]{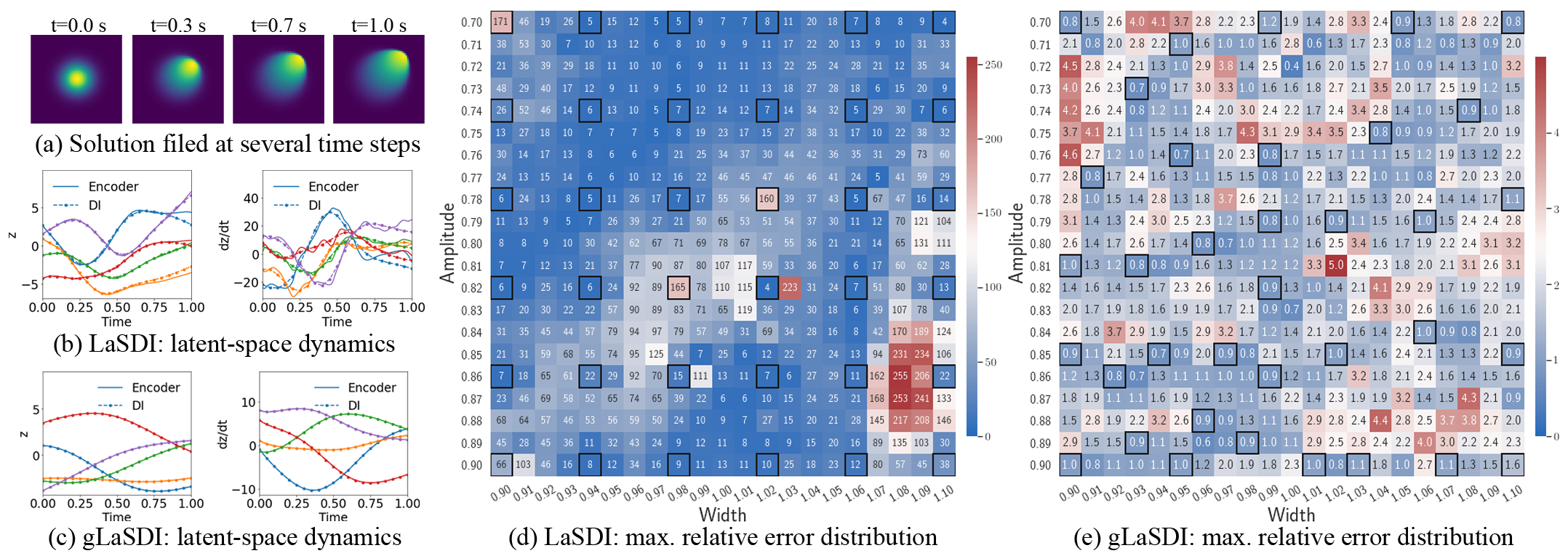}
    \caption{Comparison between LaSDI and gLaSDI with an auto-encoder of 7,200-100-5-100-7,200 and \textit{quadratic} DI models for the 2D Burgers problem. The number on each box denotes the maximum relative error of the associated parameter case. The black square boxes indicate the sampled training parameter points.}
    \label{fig.2d_burger_case1}
\end{figure}

An additional experiment and results are presented to further demonstrate the effectiveness of the proposed framework. A two-dimensional (2D) parameterized inviscid Burgers equation is considered
\begin{equation}\label{eq.2d_burger}
    \frac{\partial \mathbf{u}}{\partial t} + \mathbf{u} \cdot \nabla \mathbf{u} = \frac{1}{Re} \Delta \mathbf{u}, \quad \Omega = [-3,3]\times[-3,3], \quad t \in [0,1],
\end{equation}
with a boundary condition $\mathbf{u}(\mathbf{x},t;\boldsymbol{\mu}) = \mathbf{0}$ on $\partial\Omega$, an initial condition $\mathbf{u}(\mathbf{x},0;\boldsymbol{\mu}) = a e^{-\frac{||\mathbf{x}||^2}{w^2}}$ parameterized by $\boldsymbol{\mu} = \{a, w\}$, and a Reynolds number $Re=10,000$.
A uniform spatial discretization with $60 \times 60$ discrete points is applied. 
The first order spatial derivative and the diffusion term are approximated by the backward difference scheme and the central difference scheme, respectively. 
The full-order model solutions are obtained by solving the semi-discretized system with the implicit backward Euler time integrator and a uniform time step of $\Delta t=1/200$.
A discrete parameter space $\mathcal{D}^h$ is constituted by the parameters of the initial condition, including the width, $w \in [0.9,1.1]$, and the amplitude, $a \in [0.7,0.9]$, each with 21 evenly distributed discrete points in the respective parameter range.
The solution fields of the first velocity component at different time steps for the parameter case $(a=0.7, w=0.9)$ are shown in Fig. \ref{fig.2d_burger_case1}(a).
The auto-encoder with an architecture of 7,200-100-5-100-7,200 \XH{(latent dimension as 5)} and \textit{quadratic} DI models are employed. 
The gLaSDI training is performed until the total number of sampled parameter points reaches 36. 
A LaSDI model with the same architecture of the auto-encoder and DI models is trained using 36 predefined training points uniformly distributed in a $6 \times 6$ grid for the parameter space. 

Figs. \ref{fig.2d_burger_case1}(b-c) show that gLaSDI identifies much simpler and smoother latent-space dynamics than LaSDI, with a better agreement between the encoder and the DI predictions, which is attributed by the interactive autoencoder-DI training of gLaSDI, as illustrated in Fig. \ref{fig.glasdi} and described in Section \ref{sec:glasdi}. 
Figs. \ref{fig.2d_burger_case1}(d-e) show that gLaSDI achieves the maximum relative error of 5$\%$ in the whole parameter space, much lower than 255$\%$ of LaSDI. 
The poor accuracy of LaSDI could be caused by the deviation between the DI predicted dynamics and the encoder predicted dynamics.
It is also observed that gLaSDI tends to have denser sampling in the lower range of the parameter space. 
This demonstrates the importance of the physics-informed greedy sampling procedure.
Compared with the high-fidelity simulation based on an in-house Python code, the gLaSDI model achieves \XH{871}$\times$ speed-up.

\begin{figure}[htp]
    \centering
    \includegraphics[width=1.0\textwidth]{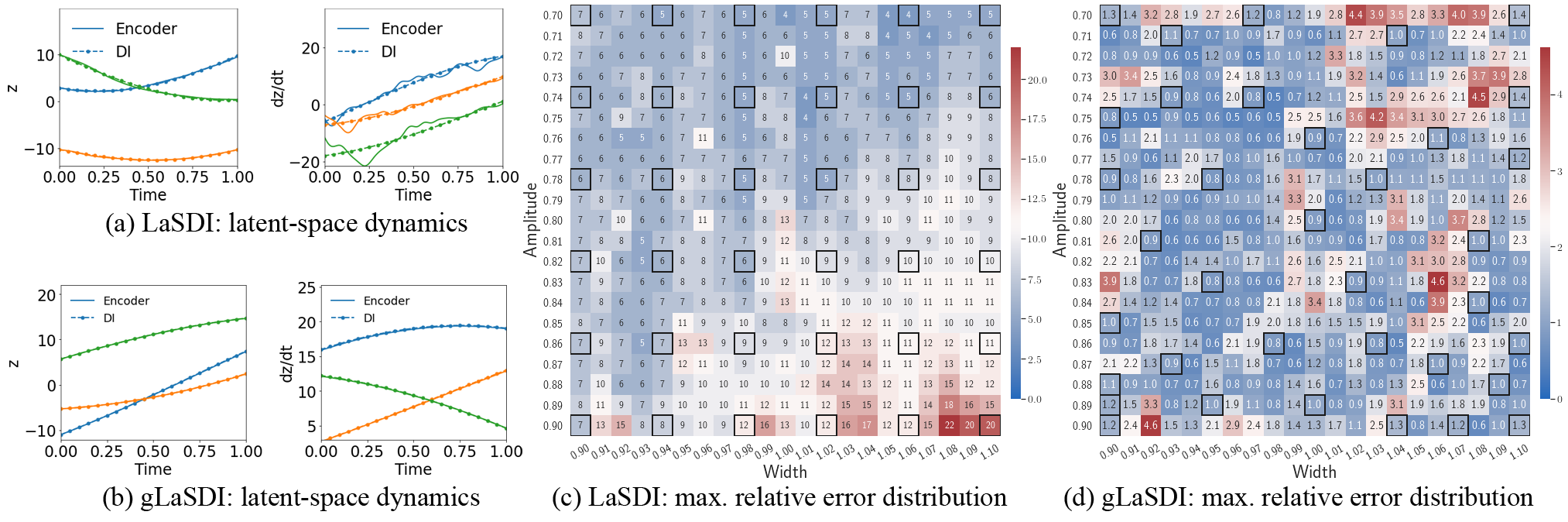}
    \caption{Comparison between LaSDI and gLaSDI with an auto-encoder of 7,200-100-3-100-7,200 and \textit{linear} DI models for the 2D Burgers problem. The number on each box denotes the maximum relative error of the associated parameter case. The black square boxes indicate the sampled training parameter points.}
    \label{fig.2d_burger_case2}
\end{figure}

\XH{When the latent dimension is reduced from 5 to 3 and the polynomial order of the DI models is reduced from quadratic to linear, gLaSDI learns simpler latent-space dynamics, as shown in Fig. \ref{fig.2d_burger_case2}(b), and maintains a high accuracy with a maximum relative error of 4.6$\%$, much lower than 22$\%$ of LaSDI, as shown in Figs. \ref{fig.2d_burger_case2}(c-d).
Furthermore, simplifying the latent-space dynamics enhances the reduced-order modeling efficiency.
Compared with the high-fidelity simulation, the gLaSDI model achieves \XH{2,658}$\times$ speed-up, which is 3.05 times the speed-up achieved by the gLaSDI model with a latent dimension of 5 and quadratic DI models.}

\section*{Acknowledgements}
This work was performed at Lawrence Livermore National Laboratory and partially funded by two LDRDs (21-FS-042 and 21-SI-006). Lawrence Livermore National Laboratory is operated by Lawrence Livermore National Security, LLC, for the U.S. Department of Energy, National Nuclear Security Administration under Contract DE-AC52-07NA27344. LLNL-CONF-835143

\printbibliography

\section*{Checklist}
\begin{enumerate}
\item For all authors...
\begin{enumerate}
  \item Do the main claims made in the abstract and introduction accurately reflect the paper's contributions and scope?
    \answerYes{}
  \item Did you describe the limitations of your work?
    \answerYes{See Section~\ref{sec:result}.}
  \item Did you discuss any potential negative societal impacts of your work?
    \answerNA{The proposed work does not involve any societal aspects.}
  \item Have you read the ethics review guidelines and ensured that your paper conforms to them?
    \answerYes{}
\end{enumerate}

\item If you are including theoretical results...
\begin{enumerate}
  \item Did you state the full set of assumptions of all theoretical results?
    \answerNA{}
  \item Did you include complete proofs of all theoretical results?
    \answerNA{}
\end{enumerate}

\item If you ran experiments...
\begin{enumerate}
  \item Did you include the code, data, and instructions needed to reproduce the main experimental results (either in the supplemental material or as a URL)?
    \answerYes{See Section~\ref{sec:result}.}
  \item Did you specify all the training details (e.g., data splits, hyperparameters, how they were chosen)?
    \answerYes{see Section~\ref{sec:result}. More details can be found in the following GitHub repository, 
    \url{https://github.com/LLNL/libROM/tree/gLaSDI/python/gLaSDI}.}
  \item Did you report error bars (e.g., with respect to the random seed after running experiments multiple times)?
    \answerNo{}
  \item Did you include the total amount of compute and the type of resources used (e.g., type of GPUs, internal cluster, or cloud provider)?
    \answerYes{See Section~\ref{sec:result}.}
\end{enumerate}

\item If you are using existing assets (e.g., code, data, models) or curating/releasing new assets...
\begin{enumerate}
  \item If your work uses existing assets, did you cite the creators?
    \answerYes{See Section~\ref{sec:result}.}
  \item Did you mention the license of the assets?
    \answerYes{See Section~\ref{sec:result}.}
  \item Did you include any new assets either in the supplemental material or as a URL?
    \answerYes{See Section~\ref{sec:result}.}
  \item Did you discuss whether and how consent was obtained from people whose data you're using/curating?
    \answerNA{The license of the used asset allows our usage of the codes and distribution.}
  \item Did you discuss whether the data you are using/curating contains personally identifiable information or offensive content?
    \answerNA{}
\end{enumerate}

\item If you used crowdsourcing or conducted research with human subjects...
\begin{enumerate}
  \item Did you include the full text of instructions given to participants and screenshots, if applicable?
    \answerNA{}
  \item Did you describe any potential participant risks, with links to Institutional Review Board (IRB) approvals, if applicable?
    \answerNA{}
  \item Did you include the estimated hourly wage paid to participants and the total amount spent on participant compensation?
    \answerNA{}
\end{enumerate}

\end{enumerate}
\newpage

\end{document}